\documentclass[10pt,twocolumn,letterpaper]{article}

\PassOptionsToPackage{table,xcdraw}{xcolor}
\usepackage{iccv}              
\usepackage[accsupp]{axessibility}
\usepackage{graphicx}
\usepackage{amsmath}
\usepackage{amssymb}
\usepackage{booktabs}
\usepackage[normalem]{ulem}
\useunder{\uline}{\ul}{}

\definecolor{iccvblue}{rgb}{0.21,0.49,0.74}
\usepackage[pagebackref,breaklinks,colorlinks,allcolors=iccvblue]{hyperref}
\usepackage{xcolor}
\definecolor{RoyalBlue}{HTML}{0072B2}
\definecolor{Vermilion}{HTML}{D55E00}
\definecolor{Bluish-Green}{HTML}{009E73}
\definecolor{RFour}{HTML}{CC79A7}

\def\paperID{****} 
\def\confName{ICCV}
\def\confYear{2025}

\title{TokensGen: Harnessing Condensed Tokens for Long Video Generation\vspace{-4mm}}

\author{
Wenqi Ouyang$^{1}$, \qquad
Zeqi Xiao$^{1}$, \qquad
Danni Yang$^{2}$, \qquad
Yifan Zhou$^{1}$,\\
\vspace{2mm}
Shuai Yang$^{3}$, \qquad\qquad
Lei Yang$^{2}$, \qquad
Jianlou Si$^{2}$, \qquad
Xingang Pan$^{1}$ 
\\
$^{1}$S-Lab, Nanyang Technological University, \qquad$^{2}$SenseTime Research, \\
$^{3}$Wangxuan Institute of Computer Technology, Peking University \\
\small{\url{https://vicky0522.github.io/tokensgen-webpage/}}\\
}

\begin{document}

\twocolumn[{
    \renewcommand\twocolumn[1][]{#1}
    \maketitle
    \centering
    \includegraphics[width=1.0\linewidth]{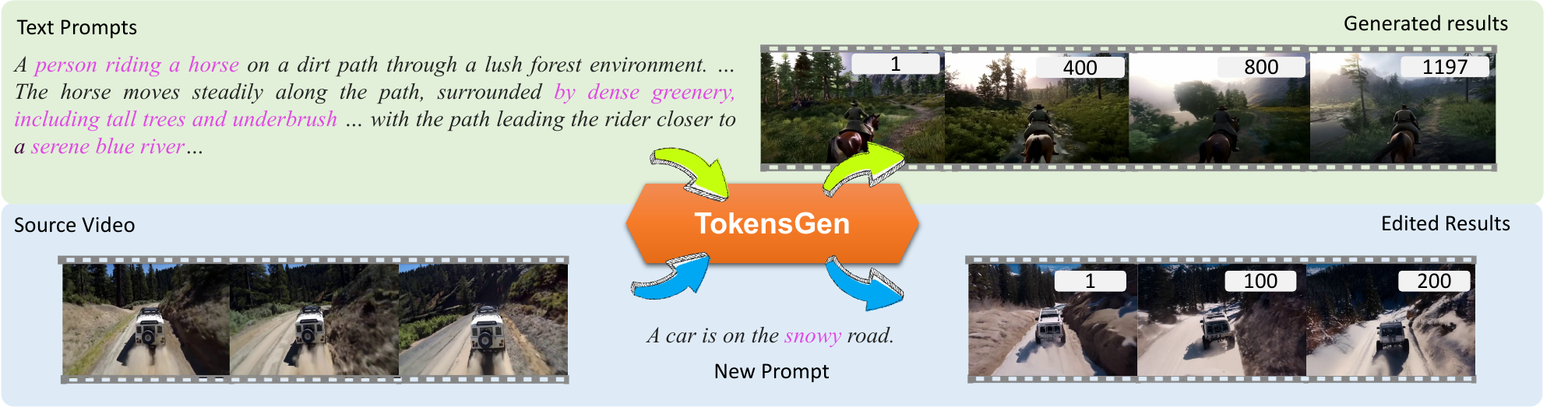}
    \captionof{figure}{Given the text prompt, TokensGen generates long videos of up to 2 minutes, maintaining consistent motions and content. Moreover, TokensGen supports zero-shot prompt-guided video-to-video editing for long videos.}
    \vspace{12pt}
     \label{fig:1_1}
}]

\begin{abstract}
Generating consistent long videos is a complex challenge: while diffusion-based generative models generate visually impressive short clips, extending them to longer durations often leads to memory bottlenecks and long-term inconsistency. In this paper, we propose TokensGen, a novel two-stage framework that leverages condensed tokens to address these issues. Our method decomposes long video generation into three core tasks: (1) inner-clip semantic control, (2) long-term consistency control, and (3) inter-clip smooth transition. 
First, we train To2V (Token-to-Video), a short video diffusion model guided by text and video tokens, with a Video Tokenizer that condenses short clips into semantically rich tokens. Second, we introduce T2To (Text-to-Token), a video token diffusion transformer that generates all tokens at once, ensuring global consistency across clips. Finally, during inference, an adaptive FIFO-Diffusion strategy seamlessly connects adjacent clips, reducing boundary artifacts and enhancing smooth transitions.
Experimental results demonstrate that our approach significantly enhances long-term temporal and content coherence without incurring prohibitive computational overhead. By leveraging condensed tokens and pre-trained short video models, our method provides a scalable, modular solution for long video generation, opening new possibilities for storytelling, cinematic production, and immersive simulations.
\end{abstract}    
\section{Introduction}
\label{sec:1}
Generating consistent and visually pleasing long videos remains a formidable challenge in the field of video generation. While diffusion-based methods excel at short video generation \cite{wang2023modelscope,wang2023lavie,chen2024videocrafter2,opensora,lin2024open,yang2024cogvideox,kong2024hunyuanvideo,kling,sora}, extending them to longer durations is constrained by computational resources, posing significant challenges.
  
Many approaches attempt to decompose long video generation into manageable sub-problems, using short video models without adding memory overhead. Tuning-free methods \cite{qiu2023freenoise,wang2023gen,tan2024videoinf,lu2024freelong,ma2025tuning,cai2024ditctrl} generate long videos using pre-trained short video models with hand-crafted techniques like noise re-scheduling, sliding window fusion, and attention manipulation, combined with multi-prompt sampling to enrich content. While ensuring high frame quality, these methods struggle with unnatural transitions due to missing long-range priors.
Auto-regressive methods \cite{chen2023seine,henschel2024streamingt2v,kim2024fifo,ruhe2024rolling,kling} and image-to-video approaches \cite{chen2023seine,xing2023dynamicrafter,blattmann2023stable,2023i2vgenxl} generate clips sequentially, achieving smooth transitions. However, they suffer from error accumulation, limited context windows, and unstable long-term controllability.

Hierarchical methods, such as MovieDreamer \cite{zhao2024moviedreamerhierarchicalgenerationcoherent}, adopt a multi-stage pipeline to address long-range challenges efficiently. They generate keyframes and then synthesize short clips guided by these keyframes, producing high-quality results. However, they mainly focus on multi-scene generation and lack strict consistency in motion and appearance across adjacent clips.

These limitations in long-term and short-term content control underscore the need for a unified, scalable solution that maintains both long-term and short-term consistency without excessive memory overhead. Therefore, we propose TokensGen, which leverages condensed video tokens to bridge short-clip generation with long-term consistency. Unlike hierarchical methods relying on keyframes generation and interpolation, or purely frame-level auto-regressive sampling, TokensGen jointly models spatial and temporal distributions for long videos through a two-stage framework, as detailed below:

a) To2V Model (Inner-clip content control): 
We employ a conditional short video generation model guided by text and video tokens to produce semantically rich yet concise video segments. Built on a powerful pre-trained backbone (CogVideoX \cite{yang2024cogvideox}), our Video Tokenizer encodes short clips into a condensed set of high-level semantic tokens. This enables robust spatial layouts and motion cues in an efficient representation space, achieving stronger per-clip semantic control than text prompts alone.

b) T2To Model (Long-term content consistency):
We train a video token diffusion transformer to generate the full set of tokens for a minute-long video from text prompts. These tokens are derived by encoding the long video clip-by-clip using the Video Tokenizer. Operating in this token space enables the T2To Model to maintain content continuity and logical coherence across clips while significantly reducing memory demands compared to raw frame modeling, preserving sufficient semantic detail for global consistency.

c)~Adaptive FIFO-Diffusion (Inter-clip temporal smoothness):
During inference, we sample long video tokens via the T2To Model and employ them to guide clip generation in the To2V Model. However, naively concatenating clips may cause boundary discontinuities, even with consistent semantic tokens. To overcome this, we propose an adaptive FIFO-Diffusion process for the To2V Model, enabling diagonal denoising of consecutive clips. This approach prevents distributional artifacts caused by naive padding or frame replication in FIFO-Diffusion \cite{kim2024fifo}, ensuring smoother transitions and improving the overall fidelity of the long video.

Compared to prior methods for long video generation, TokensGen offers several key advantages. 
First, by leveraging pre-trained short video models, it inherits strong knowledge priors and architectural designs, enabling a smooth transition from short clips to minute-long sequences without extensive re-engineering. Second, encoding long videos into condensed token representations significantly reduces computational overhead for minute-level generation. Third, because each component (To2V Model, T2To Model, and the inter-clip scheduling) operates in a clearly defined sub-task, our pipeline is highly flexible. It can seamlessly integrate with other short-term control strategies (\eg, Progressive Diffusion \cite{xie2024progressive}, Rolling Diffusion Models \cite{ruhe2024rolling}) or multi-prompt composition frameworks \cite{kling,tan2024videoinf,cai2024ditctrl}.

In summary, TokensGen offers a scalable and resource-efficient framework for generating long videos with long-term consistency and smooth transitions, as shown in \cref{fig:1_1}. By harnessing condensed tokens and powerful short video models, our approach significantly lowers the barrier to high-quality long video generation, opening new possibilities for storytelling, simulation, and beyond.
\section{Related Work}
\label{sec:2}
\textbf{Video diffusion models.}
Video diffusion models generate videos from text or image prompts. Early methods \cite{wang2023modelscope,wang2023lavie,chen2024videocrafter2,guo2023animatediff,guo2023sparsectrl,blattmann2023stable,xing2023dynamicrafter} extend U-Net-based image diffusion to the temporal domain. However, they struggle with motion dynamics and content richness due to separate spatial-temporal attention and limited temporal windows.
Recent works \cite{yang2024cogvideox,kong2024hunyuanvideo,ma2024latte,opensora,lin2024open,genmo2024mochi} enhance fidelity and consistency by integrating diffusion transformers with 3D full-attention to jointly model spatial-temporal correlations and improved text encoders for complex prompts. While effective for short videos, extending these models to long videos remains computationally prohibitive.

\noindent \textbf{Long video generation.}
Long video generation poses additional challenges for achieving content coherence, consistent dynamics, and efficient resource usage. We categorize long video generation methods into two groups: 1) those that optimize resource usage via engineering techniques or efficient model design, and 2) those that decompose long video generation into short video sub-tasks. 

\noindent \textbf{Resource usage optimization.}
Recent transformer-based methods \cite{yang2024cogvideox,kong2024hunyuanvideo,ma2024latte,opensora,lin2024open,genmo2024mochi} employ 3D-VAE to compress videos. However, as noted in CogVideoX \cite{yang2024cogvideox}, excessive compression hinders 3D-VAE convergence.
ExVideo \cite{duan2024exvideo} extends SVD \cite{blattmann2023stable} to 128 frames via small learnable parameters with low memory overhead. Pyramidal Flow Matching \cite{jin2024pyramidal} reformulates diffusion into pyramid stages, enabling efficient generation of videos up to 240 frames. While effective, these methods still face challenges in scaling to much longer durations.

\noindent \textbf{Problem decomposition via short video generation.}

\begin{itemize}
    \item \textbf{Multi-scene generation} 
Multi-shot approaches \cite{Lin2023VideoDirectorGPT,zhuang2024vlogger,Long:ECCV24,yuan2024mora} divide long videos into segments and align them under a unified narrative, conditioning video diffusion on scene-level text or styles for coherence. MovieDreamer \cite{zhao2024moviedreamerhierarchicalgenerationcoherent} employs a hierarchical pipeline to draft keyframes and refine shots. These methods emphasize storytelling and character consistency with relaxed demands on inter-clip coherence.
    \item \textbf{Tuning-free methods.}
Tuning-free methods extend short video generation to longer durations via hand-crafted designs, such as co-denoising \cite{wang2023gen,ma2025tuning}, noise re-scheduling with sliding windows \cite{qiu2023freenoise}, and attention control mechanisms \cite{lu2024freelong,cai2024ditctrl,tan2024videoinf}. Often paired with multi-prompt sampling for content richness, these methods lack long-range priors, leading to unnatural motion and appearance transitions over extended durations.
    \item \textbf{Auto-regressive methods.}
Auto-regressive methods generate long videos frame-by-frame (or clip-by-clip), ensuring temporal consistency at the cost of increased inference time. Image(clip)-to-video models \cite{blattmann2023stable,xing2023dynamicrafter,2023i2vgenxl,henschel2024streamingt2v,tian2024videotetris} generate long videos iteratively by treating the last generated frame (clip) as the initial one in the next iteration, but suffer quality degradation due to error accumulation. Loong \cite{wang2024loong} adopts an auto-regressive language model-like strategy but is limited to low resolutions ($128\times128$). Causal denoising \cite{kim2024fifo,xie2024progressive,ruhe2024rolling,yin2024slow} gradually increases noise scales for later frames, ensuring smooth transitions by referencing clearer earlier frames. Commercial tools like Kling \cite{kling} offer clip-by-clip video extension but are constrained by context windows and unstable long-term controllability.
    \item \textbf{Hierarchical methods.}
Hierarchical methods \cite{yin2023nuwa,he2022lvdm,brooks2022generating} use hierarchical frameworks to generate keyframes and perform interpolation or super-resolution. However, keyframe-only propagation increases information loss, degrading quality as stages grow. StoryDiffusion \cite{zhou2024storydiffusion} adopts a multi-stage pipeline, generating keyframes, predicting motion, and synthesizing clips guided by both. 
Different from those methods, our method models correlated spatial-temporal distribution jointly for minute-long videos via condensed tokens, utilizing the fine-grained information to guide the short clip generation to achieve more natural and consistent content across the long range.
\end{itemize}

\section{Preliminaries}
\label{sec:3}
\begin{figure}[t!]
 \centering
  \includegraphics[width=0.4\textwidth]{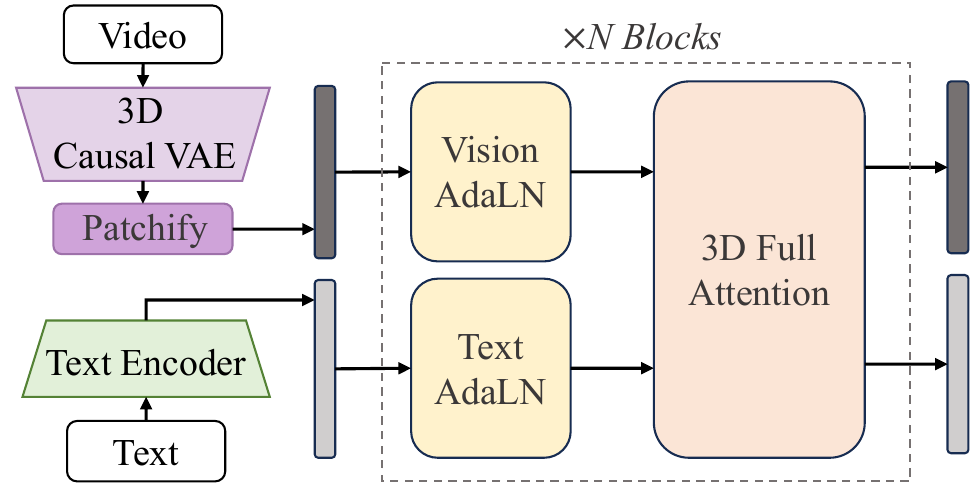}
  \caption{CogVideoX architecture.}
  \label{fig:3.1_1}
\end{figure}

\textbf{CogVideoX.}~We use the pre-trained text-to-video diffusion model CogVideoX \cite{yang2024cogvideox} as the foundation of our framework. CogVideoX employs a 3D causal VAE \cite{kingma2013auto} to compress videos into a latent space, achieving an $8 \times 8 \times 4$ compression ratio along the spatial and temporal axes.
During input processing, video latents are patchified and concatenated with text embeddings, which are then passed through Expert Transformer blocks featuring Expert Adaptive LayerNorm (AdaLN) and 3D Full Attention, as shown in Figure~\ref{fig:3.1_1}. Text and Vision Expert AdaLN separately modulate text and video features, improving alignment. To handle large motions, CogVideoX integrates 3D Text-Video Hybrid Attention with 3D Rotary Position Embedding (RoPE) \cite{su2024roformer}, effectively capturing spatial-temporal relationships.

\noindent \textbf{FIFO-Diffusion.}~FIFO-Diffusion \cite{kim2024fifo} introduces a diagonal denoising strategy to extend a pre-trained text-to-video model from $f$ frames ($f \ll M$) to generate long videos with $M$ frames. This method progressively denoises consecutive frames with increasing noise levels. Given a time step schedule $0 = \tau_{0} < \tau_{1} < ... < \tau_{f} = T$, each denoising step is defined as follows:
\begin{equation}
\begin{aligned}
[z^{1}_{\tau_{0}};...;z^{f}_{\tau_{f-1}}]=\Phi([z^{1}_{\tau_{1}};...;z^{f}_{\tau_{f}}],[\tau_{1};...;\tau_{f}],c;\epsilon_{\theta}).
\end{aligned}
\label{eq:3.2_1}
\end{equation}
The diagonal latents $\{z^{i}_{\tau_{i}}\}^{f}_{i=1}$ are stored in a queue. At each step, the foremost frame is dequeued once it reaches $\tau_{0} = 0$, while a new latent at $\tau_{f}$ is enqueued. This ensures that later frames with higher noise levels reference earlier frames with lower noise levels during denoising, maintaining temporal consistency and coherence throughout the long video generation process.
\section{TokensGen for Long Video Generation}
\label{sec:4}
\subsection{Overview}
\label{sec:4.1}
\begin{figure*}[tb]
    \centering
    \includegraphics[width=1.0\textwidth]{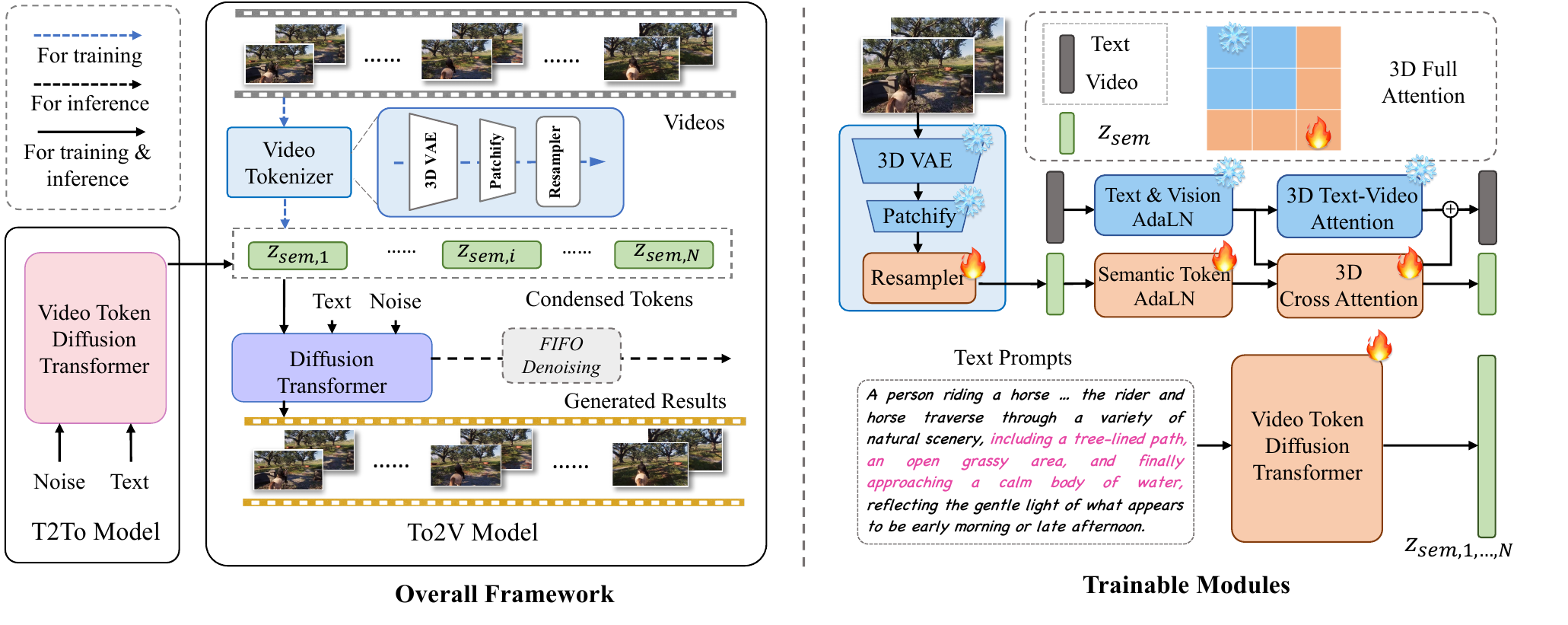}
    \caption{Overview of the model. Left: Overall Framework for TokensGen. Right: Trainable Modules.}
    \label{fig:4.1_1}
\end{figure*}
Given a text prompt, our framework generates a consistent minute-long video aligned with the prompt. It consists of two main components: To2V and T2To Models, as shown in Figure~\ref{fig:4.1_1}.
During training, we first train To2V, a conditional short video generation model, to control spatial layout and motion based on text and video prompts. A video tokenizer extracts compact semantic tokens $z_{sem}$ from short clips, which are then fed into a diffusion transformer for guided generation. Since these tokens encode richer spatial and motion information than text prompts, they provide more accurate semantic control over individual clips. For long videos, we segment them into short clips, each tokenized to produce a sequence of semantic tokens $\{z_{sem,i}\}^{N}_{i=1}$, forming a resource-efficient high-level representation of the entire video.
We then train T2To, a video token transformer, to generate these long video tokens simultaneously from text prompts, ensuring long-term content consistency across clips.
During inference, we first sample long video semantic tokens using T2To, then pass them to To2V to generate each clip. To ensure temporal consistency, we introduce an adaptive FIFO denoising strategy for diagonal denoising across clips.

\subsection{To2V Model: Inner-clip content control}
\label{sec:4.2}
We design a conditional short video generation model, To2V, guided by both text and video prompts for precise content control in short video generation. To2V builds on the pre-trained text-guided video generation model CogVideoX \cite{yang2024cogvideox} and consists of two key components:
the Video Tokenizer that encodes the input video clip into compact semantic tokens, and the Cross-Attention Branch integrated with CogVideoX that enables cross-attention between semantic tokens and noisy latents. 

\noindent \textbf{Video Tokenizer}. 
The Video Tokenizer consists of a 3D causal variational autoencoder (3D-VAE), a Patchify Module, and a Resampler, as illustrated on the right side of Figure~\ref{fig:4.1_1}. The 3D-VAE and the Patchify Module are inherited from CogVideoX with fixed weights. They process the input video into a set of tokens $z_{source}$ with the shape $f_{s}\times h_{s} \times w_{s}\times c_{s}$, where $f$, $h$, $w$, and $c$ represent the number of frames, height, width, and channels, respectively. The Resampler compresses and resamples $z_{source}$ into a more compact representation space, as illustrated in Figure~\ref{fig:4.2_1}. It comprises a learnable latent $z_{latent}$ with the shape $f_{r}\times h_{r} \times w_{r}\times c_{s}$, four blocks of the 3D Cross-Attention Module that perform cross-attention between $z_{source}$ and $z_{latent}$, and a Projector that transforms $z_{latent}$ into $z_{sem}$ with shape $f_{r}\times h_{r} \times w_{r}\times c_{r}$, where $f_{r}<f_{s}, h_{r}<h_{s}, w_{r}<w_{s}, c_{r}<c_{s}$. 
The semantic tokens $z_{sem}$ encoded by the Video Tokenizer encapsulate high-level spatial layouts and motion information from the input video while maintaining a more compact size compared to the original video.

\noindent \textbf{Cross-Attention Branch}. 
To effectively incorporate these semantic tokens with CogVideoX, we add a separate Cross-Attention Branch to handle the newly added semantic conditions. This branch consists of a Semantic Token Adaptive LayerNorm (Sem AdaLN) and a 3D Cross-Attention Module, as depicted on the right side of Figure~\ref{fig:4.1_1}. The process is as follows: 
\begin{figure}[!tb]
    \centering
    \includegraphics[width=0.5\textwidth]{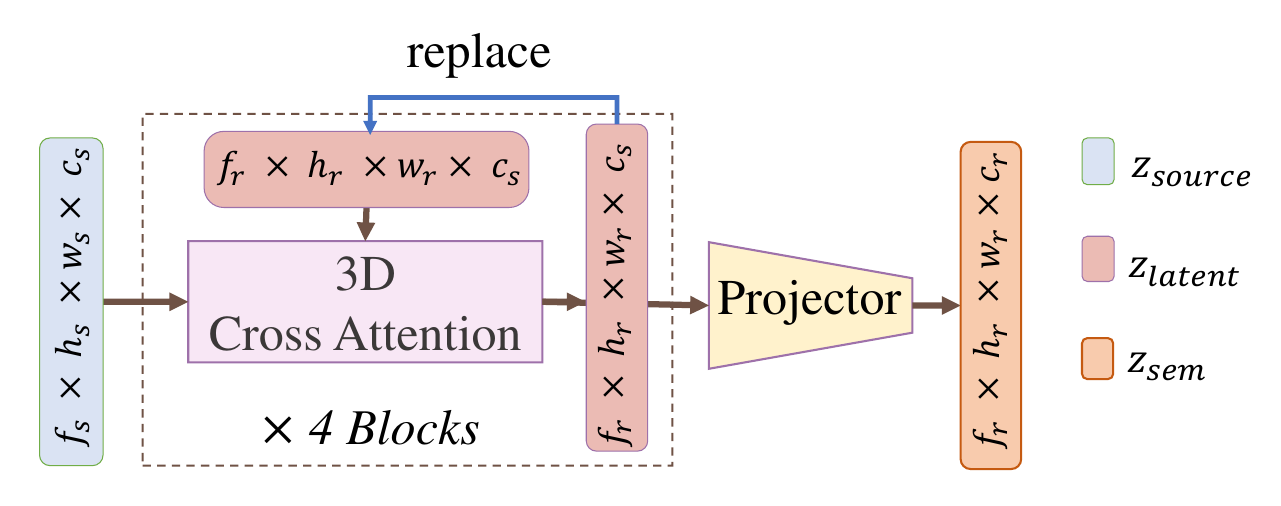}
    \caption{The architecture of the Resampler.}
    \label{fig:4.2_1}
\end{figure}

\begin{itemize}
    \item \textbf{Back projection}: The semantic tokens $z_{sem}$ from the Video Tokenizer are back-projected to match the number of channels of the combined text-video embeddings $\mathbf{Z_{tv}} = [\mathbf{Z_{text}};\mathbf{Z_{video}}]$.
    \item \textbf{Concatenation}: These back-projected semantic tokens are concatenated with the text-video embeddings.
    \item \textbf{Modulation}: Similar to the Text and Vision AdaLN, the Sem AdaLN modulates the semantic condition embeddings to ensure better feature alignment.
    \item \textbf{Attention}: The modulated embeddings are passed to the 3D Text-Video Attention and the 3D Cross Attention Module to perform 3D full attention on the combined embeddings. Given the combined embeddings $[\mathbf{Z_{text}};\mathbf{Z_{video}};\mathbf{Z_{sem}}]$, the output attention results $[\mathbf{Z^{'}_{text}};\mathbf{Z^{'}_{video}};\mathbf{Z^{'}_{sem}}]$ are represented as follows:
\begin{equation}
\begin{aligned}
&\mathbf{Z_{tv}} = [\mathbf{Z_{text}};\mathbf{Z_{video}}] \\
\mathbf{Q} = \mathbf{Z_{tv}}&\mathbf{W_{q}}, \mathbf{K} = \mathbf{Z_{tv}W_{k}}, \mathbf{V} = \mathbf{Z_{tv}W_{v}} \\
\mathbf{Q_{s}} = \mathbf{Z_{sem}}&\mathbf{W^{c}_{q}}, \mathbf{K_{s}} = \mathbf{Z_{sem}W^{c}_{k}}, \mathbf{V_{s}} = \mathbf{Z_{sem}W^{c}_{v}} \\
\mathbf{Q_{tv}} = \mathbf{Z_{tv}}&\mathbf{W^{c}_{q}}, \mathbf{K_{tv}} = \mathbf{Z_{tv}W^{c}_{k}}, \mathbf{V_{tv}} = \mathbf{Z_{tv}W^{c}_{v}} \\
[\mathbf{Z'_{text}};\mathbf{Z'_{video}}] &= {\rm Attn}(\mathbf{Q,K,V}) + {\rm Attn}(\mathbf{Q_{tv},K_{s},V_{s}}) \\
= {\rm softmax}&(\frac{\mathbf{Q}(\mathbf{K})^{T}}{\sqrt{d}})\mathbf{V} + {\rm softmax}(\frac{\mathbf{Q_{tv}}(\mathbf{K_{s}})^{T}}{\sqrt{d}})\mathbf{V_{s}}, \\
\mathbf{Z'_{sem}} &= {\rm Attn}(\mathbf{Q_{s},[K_{tv};K_{s}],[V_{tv};V_{s}]}) \\
&= {\rm softmax}(\frac{\mathbf{Q_{s}}(\mathbf{[K_{tv};K_{s}]})^{T}}{\sqrt{d}})\mathbf{[V_{tv};V_{s}]},
  \end{aligned}
  \nonumber
  \label{eq:4.1_1}
\end{equation}
where $\mathbf{W_{q}}$, $\mathbf{W_{k}}$, $\mathbf{W_{v}}$ are fixed parameters of the 3D Text-Video Attention Module inherited from CogVideoX, $\mathbf{Q}$, $\mathbf{K}$, $\mathbf{V}$ are the corresponding query, key, and value matrices. $\mathbf{W^{c}_{q}}$, $\mathbf{W^{c}_{k}}$, $\mathbf{W^{c}_{v}}$ are trainable parameters of the 3D Cross-Attention Module,  $\mathbf{Q_{s}}$, $\mathbf{K_{s}}$, $\mathbf{V_{s}}$ are the query, key, and value matrices for $\mathbf{Z_{sem}}$, while $\mathbf{Q_{tv}}$, $\mathbf{K_{tv}}$, $\mathbf{V_{tv}}$ are the query, key, and value matrices for the combined text and video embeddings $\mathbf{Z_{tv}}$. 
\end{itemize}
\begin{figure*}[!tb]
    \centering
    \includegraphics[width=1.0\textwidth]{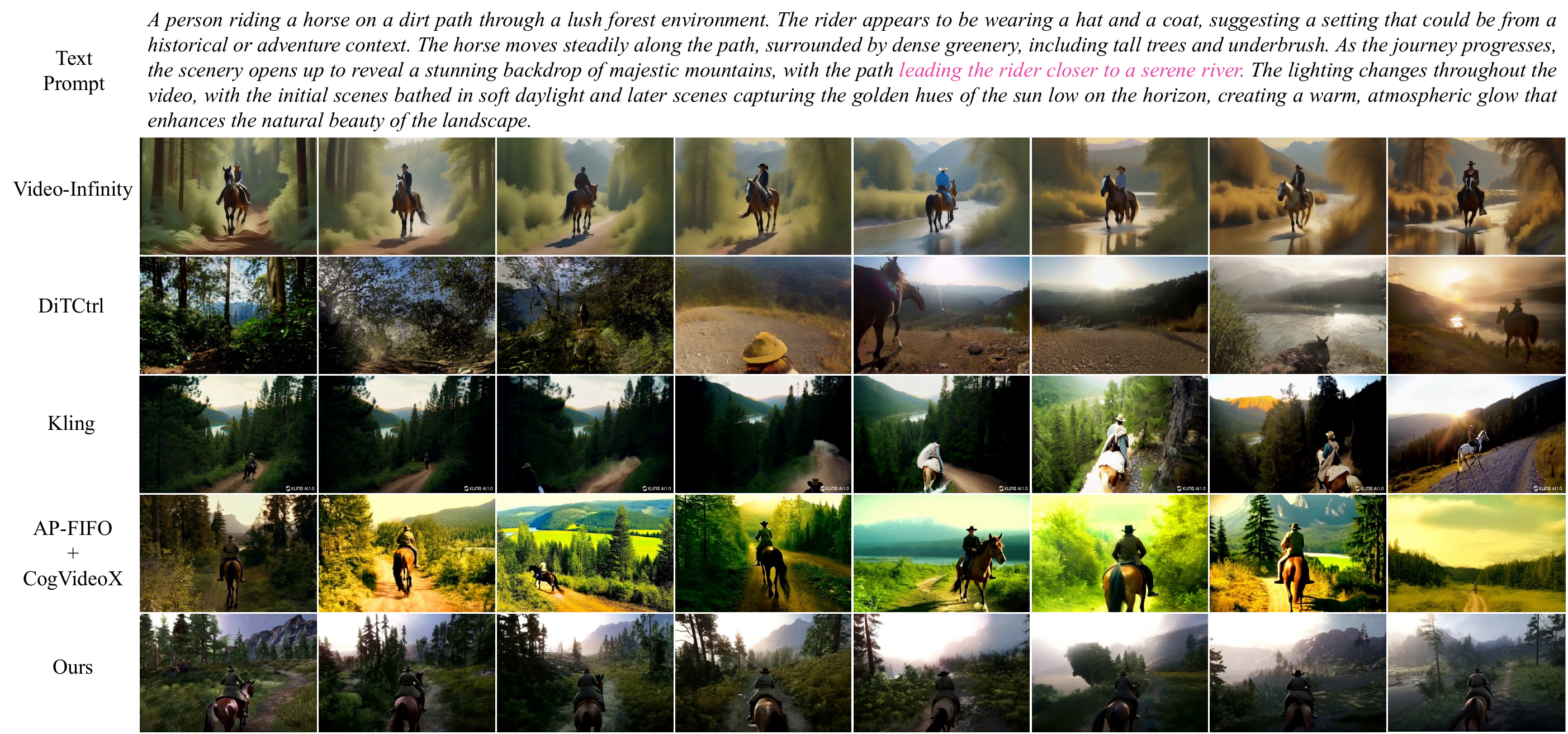}
    \caption{The qualitative comparison. 
    We recommend readers refer to our webpage for video comparisons.
    }
    \label{fig:5_1}
\end{figure*}

\subsection{T2To Model: Long-term content consistency}
\label{sec:4.3}
To learn long-term content and logic knowledge across the minute-long video, we design a video token transformer, the T2To Model, to generate the semantic tokens $\{z_{sem,i}\}^{N}_{i=1}$ representing the whole long video given the input text prompt. We adopt the same model structure and training strategy of CogVideoX \cite{yang2024cogvideox} for the T2To Model, except for the following modifications:

\begin{itemize}
    \item The model aims to generate $\{z_{sem,i}\}^{N}_{i=1}$ with the shape $(Nf_{r})\times h_{r}\times w_{r}\times c_{r}$. The total number of tokens is $(Nf_{r})\times h_{r}\times w_{r}$.
    \item Since the number of tokens along the temporal dimension is much larger than the spatial dimensions, for 3D-RoPE, we reallocate the hidden state channel for height, width, and temporal dimension as about $10\%, 10\%, 80\%$.
\end{itemize}

\subsection{Inter-clip temporal consistency}
\label{sec:4.4}
If each clip is denoised separately with a corresponding semantic token $z_{sem,i}$, the model will generate a group of discontinuous clips. To achieve temporal continuity, we apply the FIFO-denoising strategy during the inference stage. Specifically, we adopt latent partitioning and lookahead denoising, like the original FIFO. However, to maintain a queue with sufficient frames at the start of denoising, FIFO pads the positions ahead of the first clip with noise-augmented first frame replications. We observe that this approach introduces artifacts in our settings, as the replicated frames deviate from the intrinsic distribution of the training domain for the video diffusion model. To address this, we propose an improved version of FIFO, named adaptive-FIFO, which incorporates an adaptive padding strategy at the beginning of the denoising process. For a latent partition containing fewer than $f_{s}$ frames, we denoise all frames together and update them simultaneously. For a partition with exactly $f_{s}$ frames, we employ lookahead denoising: the frames are denoised together, but only the noisier frames in the latter portion are updated. By better aligning the initial padding with the model's learned distribution and ensuring continuity in partially filled partitions, this approach yields smoother transitions across clips and better frame quality.

\subsection{Training strategy}
\label{sec:4.5}
For To2V Model, we fix the weights of the pre-trained modules of the base model, train the Resampler of the Video Tokenizer and the Cross-Attention Branch. For T2To Model, we initialize the model with the weights of the base model and train all the modules. 

We adopt similar training strategies with CogVideoX \cite{yang2024cogvideox}, including Multi-Resolution Frame Pack and Explicit Uniform Sampling. 
For T2To Model, we pack videos with different time duration into the same batch, and apply an attention mask indicating the valid frames, as well as the attention mask for loss calculation, to ensure attention module focus on the right area of the input noisy latents, an approach similar with Patch'n Pack \cite{DBLP:conf/nips/0001MDHMCSPGAOP23}.
For both To2V and T2To Model, we employ the explicit uniform sampling strategy for sampling timesteps.
\setlength{\textfloatsep}{1mm}

\begin{table*}[tb]
\centering
\setlength{\tabcolsep}{3pt}
\caption{Quantitative evaluation of comparison study.}
\label{tab:5.3.1}
\resizebox{1.0\linewidth}{!}{
\begin{tabular}{@{}ccccccccc@{}}
\toprule
                 & \multicolumn{6}{c}{VBench}                                                                                                                                                                                                                                                                                                                                                                                 & \multicolumn{2}{c}{Human Study}                                                                                                                                                    \\ \midrule
Models           & \begin{tabular}[c]{@{}c@{}}Subject \\ Consistency\end{tabular} & \begin{tabular}[c]{@{}c@{}}Background\\ Consistency\end{tabular} & \begin{tabular}[c]{@{}c@{}}Temporal\\ Flickering\end{tabular} & \begin{tabular}[c]{@{}c@{}}Motion\\ Smoothness\end{tabular} & \begin{tabular}[c]{@{}c@{}}Imaging\\ Quality\end{tabular} & \multicolumn{1}{c|}{\begin{tabular}[c]{@{}c@{}}Dynamic\\ Degree\end{tabular}} & \multicolumn{1}{c}{\begin{tabular}[c]{@{}c@{}}Text-Visual \\ Alignment\end{tabular}} & \multicolumn{1}{c}{\begin{tabular}[c]{@{}c@{}}Motion \& Content \\ Consistency\end{tabular}} \\ \midrule
Video-Infinity   & 81.80                                                          & 90.56                                                            & \textbf{96.66}                                                         & 97.65                                                       & 61.58                                                     & \multicolumn{1}{c|}{31.0}                                                    &                                                                                  0.0\%    &    0.69\%                                                                                         \\
DiTCtrl          & 76.76                                                          & 87.96                                                            & \underline{95.52}                                                         & \underline{97.78}                                                       & 59.26                                                     & \multicolumn{1}{c|}{75.0}                                                    &                                                                                  0.0\%    &    1.0\%                                                                                         \\
AP-FIFO+CogVideoX & \textbf{86.22}                                                          & \textbf{92.89}                                                            & 94.78                                                         & 97.48                                                       & \textbf{64.10}                                                     & \multicolumn{1}{c|}{\underline{78.57}}                                                   &                                                                                  24.31\%    &   22.57\%                                                                                          \\ \midrule
Ours             & \underline{84.57}                                                          & \underline{92.2}                                                            & 95.41                                                         & \textbf{98.08}                                                       & \underline{63.31}                                                     & \multicolumn{1}{c|}{\textbf{78.95}}                                                   &                                                                                  \textbf{75.69\%}    &   \textbf{75.74\%}                                                                                         \\ \bottomrule
\end{tabular}}
\end{table*}

\begin{table}[tb]
\centering
\caption{Ablation study on ways of incorporating video conditions.}
\label{tab:5.3.2}
\resizebox{0.9\linewidth}{!}{\begin{tabular}{@{}cccc@{}}
\toprule
Methods        & MSE$\downarrow$     & SSIM$\uparrow$ & temporal flckering$\uparrow$ \\ \midrule
SR             & \textbf{1.12e-2} & \underline{0.60} & 97.10              \\
\begin{tabular}[c]{@{}c@{}}4x8x12\\ (w/o projection)\end{tabular} & \underline{1.24e-2} & \textbf{0.62} & \textbf{97.58}              \\
4x5x7          & 2.84e-2 & 0.49 & \underline{97.57}              \\
13x5x7         & 1.54e-2 & 0.58 & 97.54              \\
4x8x12         & \underline{1.24e-2} & \textbf{0.62} & \textbf{97.58}              \\ \bottomrule
\end{tabular}}
\end{table}
\section{Experimental Results}
\label{sec:5}
\subsection{Implementation Details}
\label{sec:5.1}
\noindent \textbf{Model Architecture.}
We employ CogVideoX-5B~\cite{yang2024cogvideox} as the base model for both To2V and T2To. In To2V, the input tokens $z_{source}$ have the shape \(13\times30\times45\times3072\). We observed that T2To struggles to converge when $c_{r}$ is large (\eg, comparable to $c_{s}$), so we set the compressed semantic tokens $z_{sem}$ to have dimensions \(4\times8\times12\times16\). For the Projector in the Resampler, we observed that a linear projection via PCA \cite{doi:10.1080/14786440109462720} effectively reduces the channel dimension without sacrificing information, as further analyzed in \cref{sec:5.3}. 
Compared to the original latent shape \(13\times60\times90\times16\), we achieve a compression ratio of approximately \(3\times8\times8\). 
Thus, we first train To2V without the channel projection and then apply PCA to the Resampler's output embeddings on 300 samples to obtain the transformation matrix.
In T2To, we set the maximum number of chunks \(N=24\). Each chunk contains $49$ frames, allowing our model to process videos up to \(24\times 49 = 1176\) frames.

\noindent\textbf{Dataset.}
We use the MiraData dataset \cite{ju2024miradatalargescalevideodataset}, comprising long videos with structured captions. We first collect 56k videos, using their \emph{dense captions} for training. 
For To2V Model, we randomly sample 49-frame video clips at 10~fps from these long videos as training targets. 
For T2To Model, we select a filtered subset of around 16k high-quality videos that are at least one minute long, primarily consisting of gameplay footage and natural landscape. We filter out videos with abrupt scene changes via PySceneDetect \cite{pyscenedetect} and human evaluations.
This subset ensures consistency within long videos for the training of T2To Model. 

\noindent \textbf{Training Details.}
We adopt a progressive learning strategy for both T2To and To2V Model. For To2V Model, we first train on smaller token shapes \((4\times 5\times 7\times 3072)\) for 1,200 iterations, using a batch size of 72 and a learning rate of \(1\times 10^{-3}\). We then transition to the full token resolution \((4\times 8\times 12\times 3072)\) for 2,600 more iterations, initializing from the previously trained model. For T2To Model, we begin with shorter videos (\(N=3\) chunks of 49 frames each) for 1,200 iterations, using a learning rate of \(1\times 10^{-3}\). Next, we expand to long videos with up to \(N=24\) chunks, training for 5,000 iterations, with a learning rate of \(3\times 10^{-4}\) and a batch size of 105.
This progressive training helps the model converge faster for more complex, extended video generation.

\subsection{Baseline comparisons}
\label{sec:5.2}
\noindent \textbf{Qualitative comparisons}
We evaluate our approach against several recent multi-prompt long video generation methods, including Video-Infinity~\cite{tan2024videoinf}, DiTCtrl~\cite{cai2024ditctrl}, and Kling~\cite{kling}, as well as a baseline that adopts FIFO-Diffusion \cite{kim2024fifo} on CogVideoX with our adaptive-padding strategy. For multi-prompt methods, we use GPT-4o \cite{gpt-4o} to split the prompt into 24 segments, which are used for guiding each clip generation. FIFO and ours use the same text prompt, abbreviated as: \emph{``a person riding a horse along a path leading to a serene river.''} The results are shown in \cref{fig:5_1}.
Video-Infinity produces transitions primarily through background changes while failing to capture meaningful foreground motion. The person and the horse remain essentially static within each clip, resulting in low engagement and narrative drift over longer durations. DiTCtrl demonstrates intermittently aligned keyframes but struggles to maintain smooth transitions between clips, leading to abrupt scene shifts and a disjointed storyline. Kling generates visually consistent frames but exhibits erratic motions (\textit{e.g.}, the subject abruptly reversing direction) and occasional discontinuities in scene composition. Such artifacts disrupt the viewing experience and deviate from the intended story arc. For FIFO (with adaptive padding on CogVideoX), we observe progressive over-saturation and abrupt changes in appearance or color schemes as the video extends. These issues are especially pronounced when generating complex scenes over hundreds of frames. By contrast, our method delivers smoother motion transitions and subject representation, adherent to the prompt throughout the entire minute-long sequence. More comparison results are included in the \cref{supp:add_comparison}.
\begin{figure*}[!tb]
    \centering
    \includegraphics[width=1.0\textwidth]{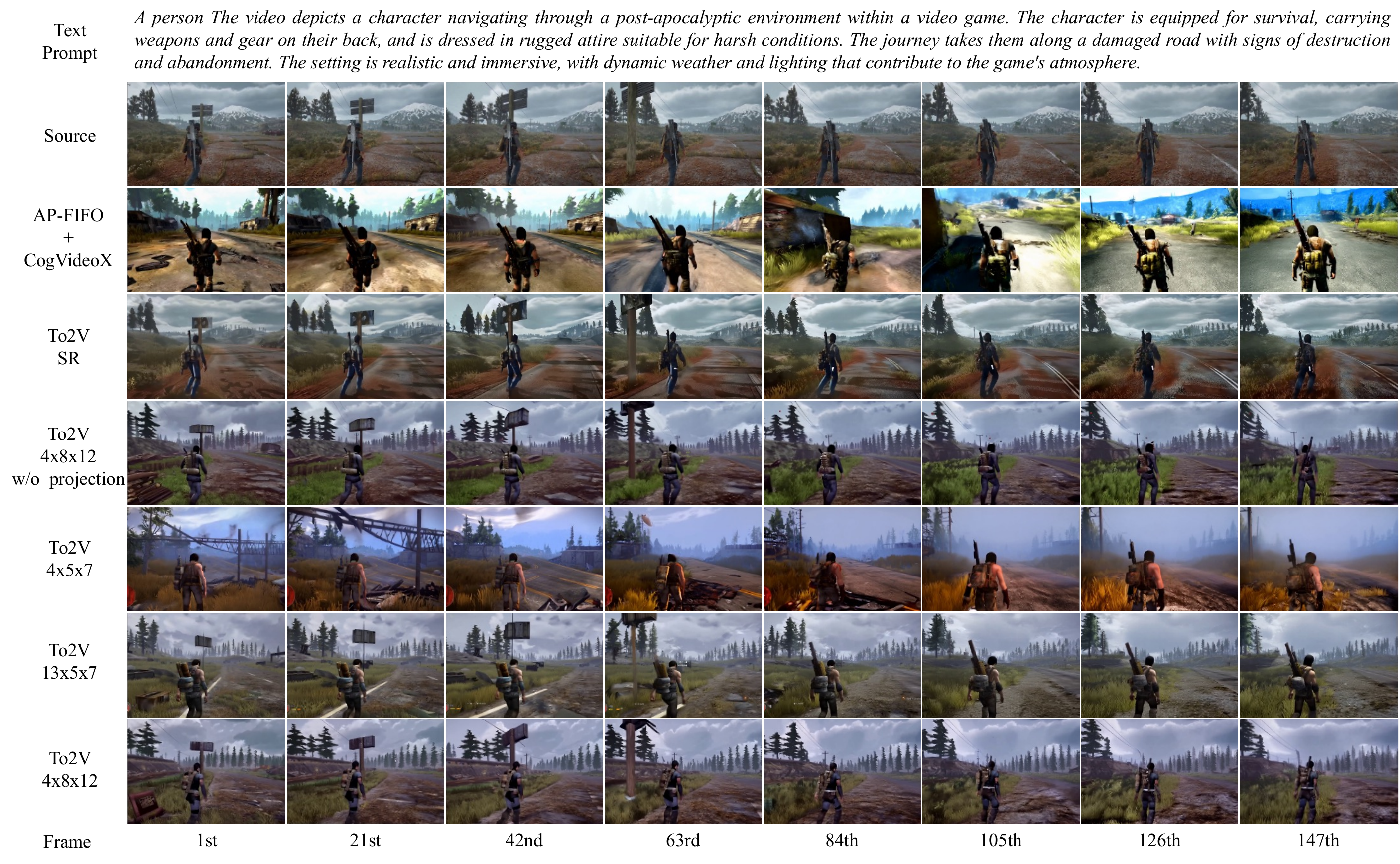}
    \caption{Ablation study on methods of incorporating video conditions.}
    \label{fig:5_2}
\end{figure*}
\begin{figure*}[!tb]
    \centering
    \includegraphics[width=1.0\textwidth]{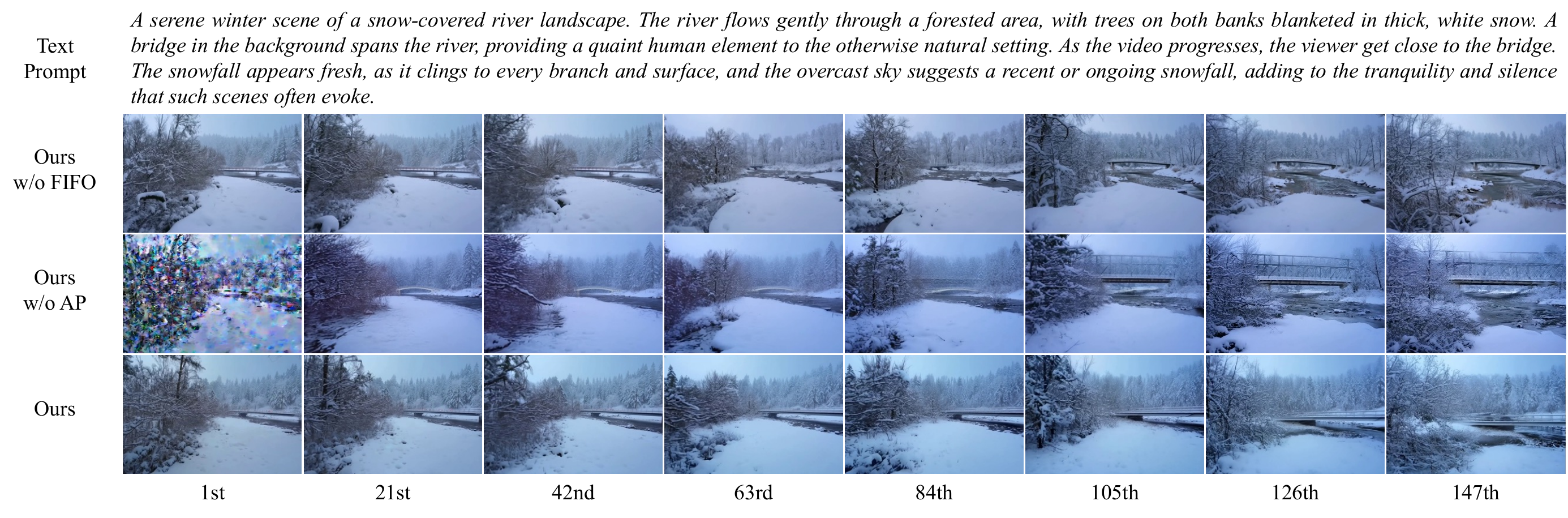}
    \caption{Ablation study FIFO.}
    \label{fig:5_3}
\end{figure*}

\noindent \textbf{Quantitative comparisons.}
We conduct a quantitative comparison study on 100 prompts randomly selected from the MiraData~\cite{ju2024miradatalargescalevideodataset} test set. As reported in \cref{tab:5.3.1}, our approach achieves the highest scores on VBench~\cite{huang2023vbench} for both \emph{Motion Smoothness} and \emph{Dynamic Degree}. We observe that certain metrics in VBench (\eg, \emph{Subject and Background Consistency, and Temporal Flickering}) may assign higher scores to less dynamic videos, motivating us to conduct a user study for a more comprehensive evaluation.
For the user study, we generate 12 video results for each method, with each video lasting between one and two minutes, covering categories such as humans, cars, and natural scenes. All the resulting videos are included on our webpage.
To ensure unbiased feedback, videos are randomly shuffled and presented to $24$ participants. They evaluate each video on two criteria: \emph{text-visual alignment} and \emph{motion \& content consistency}. As shown on the right side of \cref{tab:5.3.1}, our method consistently outperforms others in both dimensions, reflecting its strong long-term control capabilities. These results demonstrate that our approach effectively maintains semantic alignment with the textual prompt while preserving smooth motion and stable content over extended sequences. Additional results and details are provided in the \cref{supp:comparison_study}. 

\subsection{Ablation studies}
\label{sec:5.3}
\noindent \textbf{Ablation on video conditions.}
We investigate various strategies for incorporating video conditions into the To2V Model: (1) the condensed token shape, (2) with or without channel projection, and (3) a super-resolution-based approach. Specifically, we experiment with three token shapes $(4\times5\times7,13\times5\times7,4\times8\times12)$, train models with and without the Projector module, and compare them against a super-resolution setting (where the low-resolution video directly serves as the condition). We also include a baseline using FIFO-Diffusion to illustrate its potential shortcomings without the conditioning process.
As shown in Figure~\ref{fig:5_2}, the FIFO-based approach often produces inconsistent foreground and background visuals, underscoring the difficulty of preserving spatial-temporal coherence from purely latent-level guidance. Meanwhile, the super-resolution method tends to duplicate low-level color and texture cues from the source, failing to capture higher-level semantics, leading to less meaningful renderings. Comparing models with and without the Projector, we observe similar performances, demonstrating that our PCA-based projection provides a lightweight yet effective means of dimensionality reduction without sacrificing image quality. Concerning the shape of the condensed tokens, the smallest variant ($4\times5\times7$) struggles to preserve essential layout and motion patterns, resulting in less accurate re-creations of the source video. Increasing the token's temporal or spatial resolution ($13\times5\times7,4\times8\times12$) significantly improves alignment and maintains better semantic fidelity. Among these, ($4\times8\times12$) achieves the most favorable balance of fine-grained detail and computational efficiency, as quantitatively confirmed in \cref{tab:5.3.2}. Overall, these ablation studies demonstrate that our token representation, combined with optional PCA-based projection, offers a robust and effective pathway to incorporate video conditions in To2V Model.


\noindent \textbf{Ablation on FIFO.}
We further examine the influence of FIFO-Diffusion and our adaptive padding technique by comparing three variants: (1) omitting FIFO entirely, (2) using FIFO without adaptive padding, and (3) our full approach, incorporating both FIFO and adaptive padding.
As illustrated in \cref{fig:5_3}, disabling FIFO leads to abrupt scene changes between consecutive clips, producing visually inconsistent transitions where subjects may teleport or backgrounds shift abruptly. Meanwhile, removing adaptive padding induces severe artifacts in the initial frames of the video, as the model relies on duplicated frames that deviate from the training distribution. These artifacts propagate into subsequent frames, degrading overall quality. In contrast, our adaptive padding strategy aligns the padding frames with the model's distribution, preventing unnatural discontinuities at clip boundaries.

\subsection{Long Video Editing}
\begin{figure*}[!htbp]
    \centering
    \includegraphics[width=1.0\textwidth]{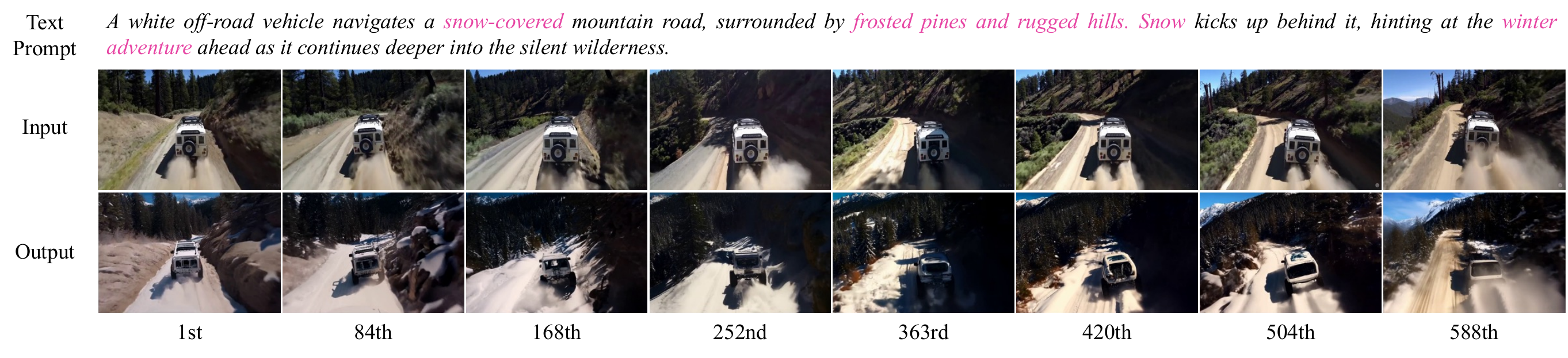}
    \caption{Long Video Editing.}
    \label{fig:editing}
\end{figure*}

Beyond generating entirely novel content, our method readily adapts to various long video editing scenarios. Specifically, the To2V Model's capacity to integrate textual prompts with source video data allows for transformations that preserve the essential structure of the original footage while injecting new semantics. We directly combine the target text prompt and the source video as input conditions to generate the edited long video, as shown in \cref{fig:editing}. For more results, please refer to our webpage.

\section{Conclusion and Discussion}
\label{sec:6}
We introduce TokensGen, a two-stage framework that addresses key challenges in long video generation, controlling per-clip semantics, ensuring long-term coherence, and achieving smooth transitions. The To2V Model generates short clips guided by text and video prompts, capturing fine-grained motion and content. The T2To Model transformer then leverages condensed semantic tokens to preserve long-term consistency across clips. Finally, our adaptive FIFO-Diffusion strategy overcomes boundary artifacts by maintaining temporal continuity. 
This pipeline efficiently scales pre-trained short video models to longer videos, enabling a scalable, flexible, and resource-efficient approach to long video generation.

Despite the effectiveness of TokensGen in maintaining long-range consistency, it does not preserve all fine-grained details. Focusing on high-level semantics, tokens may cause gradual variations in foreground objects over extended sequences (detailed in the supplementary). In complex scenes, their capacity to capture intricate spatial-temporal cues may be insufficient, requiring fine-grained tokenization and stronger short-term consistency strategies beyond tuning-free FIFO. 
Our framework is tested on a limited dataset of gameplay and landscape videos, but is scalable to larger datasets for broader applications. In future work, exploring multi-scale tokenization or hybrid representations could bolster fine-grained controllability, retaining subtle attributes while preserving the scalability and resource efficiency.

\section*{Acknowledgements}
\vspace{-0.5em}
\noindent
This research is supported by NTU SUG-NAP and is also supported under the RIE2020 Industry Alignment Fund – Industry Collaboration Projects (IAFICP) Funding Initiative, as well as cash and in-kind contribution from the industry partner(s).

{
    \small
    \bibliographystyle{ieeenat_fullname}
    \bibliography{main}
}

\clearpage
\onecolumn
\section*{Appendix}
\appendix



%
\noindent\textbf{Overview.} The appendix includes sections as follows:

\begin{quote}
\begin{itemize}
    \item Details of Comparison Study
    (\cref{supp:comparison_study}).

    \item Additional Comparisons
    (\cref{supp:add_comparison}).

    \item Limitations and Discussions
    (\cref{supp:limitations}).

    \item Additional Visual Results
    (\cref{supp:add_results}).
\end{itemize}
\end{quote}


\section{Details of Comparison Study}
\label{supp:comparison_study}
\subsection{Prompt Splitting}
When comparing our approach with multi-prompt methods such as Video-Infinity \cite{tan2024videoinf}, DiTCtrl \cite{cai2024ditctrl}, and Kling \cite{kling}, we first divide the input text prompt into several chunks to guide the generation of individual clips. Specifically, for DiTCtrl and Kling, we employ GPT-4o \cite{gpt-4o} to split the provided prompt into 24 chunks for a 2-minute-long video or 13 chunks for a 1-minute-long video, using the following instructions:
\vspace{0.5em}

\textit{Please split the prompt depicting a video into 24 separate prompts, each depicting a specific range of the duration of the video in order, and each should have the same style and length as the original prompt. Each prompt should be strictly aligned with the original prompt; if additional content is added, it should also be aligned with the scenery of the original prompt. Each prompt should occupy one line. Please do not insert a blank line between two prompts.}

\textit{The output format is as follows:}

\textit{$<$split prompt 1$>$}

\textit{$<$split prompt 2$>$}

\textit{$<$split prompt 3$>$}

\textit{...}

\textit{$<$split prompt 24$>$}
~\\
\textit{The prompt needs to be split is: }

\textit{$<$paste the input text prompt here$>$}
\vspace{0.5em}

For Video-Infinity, which is built on VideoCrafter2 \cite{chen2024videocrafter2} supporting text prompts of up to 77 tokens, we utilize its ability to perform parallel inference across 8 GPUs. To efficiently split text prompts for this method, we provide GPT-4o \cite{gpt-4o} with the following instructions:
\vspace{0.5em}

\textit{Please split the prompt depicting a video into 8 separate prompts, each depicting a specific range of the duration of the video in order, and each should have the same style as the original prompt. Each prompt should be strictly aligned with the original prompt, if additional content is added, it should also be aligned with the scenery of the original prompt. Each prompt should have fewer than 55 words. Please do not insert a blank line between two prompts.}

\textit{The output format is as follows:}

\textit{$<$split prompt 1$>$}

\textit{$<$split prompt 2$>$}

\textit{$<$split prompt 3$>$}

\textit{...}

\textit{$<$split prompt 8$>$}
~\\
\textit{The prompt needs to be split is: }

\textit{$<$paste the input text prompt here$>$}
\vspace{0.5em}

Although we provided detailed instructions, we observed that this task remains highly challenging. GPT-4o often generates split prompts where each segment contains words with a different total number than the original prompt, deviating from the intended style and length. To ensure reproducibility and facilitate comparison, we include all the text prompts along with their corresponding split versions used in the study in the accompanying supplementary material.

\subsection{User Study}
We conduct a user study to further evaluate the effectiveness of our method. Test prompts are collected from MiraData \cite{ju2024miradatalargescalevideodataset}. For multi-prompt methods, we split the text prompts using the approaches described in the previous section. For A-FIFO+CogVideoX, the same input text prompt as our method is used.
In total, we generate 12 video results for each method, with each video ranging from 1 to 2 minutes in length. The test categories include humans, cars, and natural scenes. All videos used in the user study are displayed on our webpage.
To ensure an unbiased evaluation, the results are randomly shuffled and displayed to $24$ participants. Participants are asked to evaluate the videos based on two aspects: text-visual alignment and motion and content consistency.  Questions for each aspect are as follows:
\begin{quote}
\begin{itemize}
    \item \textit{Which one best aligned the given text?}
    
    \item \textit{Which one keeps the best motion and content consistency in the long-range? For example, the video does not demonstrate scene disjoint, unreasonable content, or obvious quality degradation.}
\end{itemize}
\end{quote}

Our method achieves the best performance across all aspects of the human evaluations, as presented in our main paper. These results highlight the superior long-term control capability of our proposed method, effectively demonstrating its ability to maintain text-visual alignment and ensure motion and content consistency over extended video durations.

\section{Additional Comparisons and Analysis}
\label{supp:add_comparison}
\begin{figure*}[!htbp]
  \centering
  \includegraphics[width=1.0\linewidth]{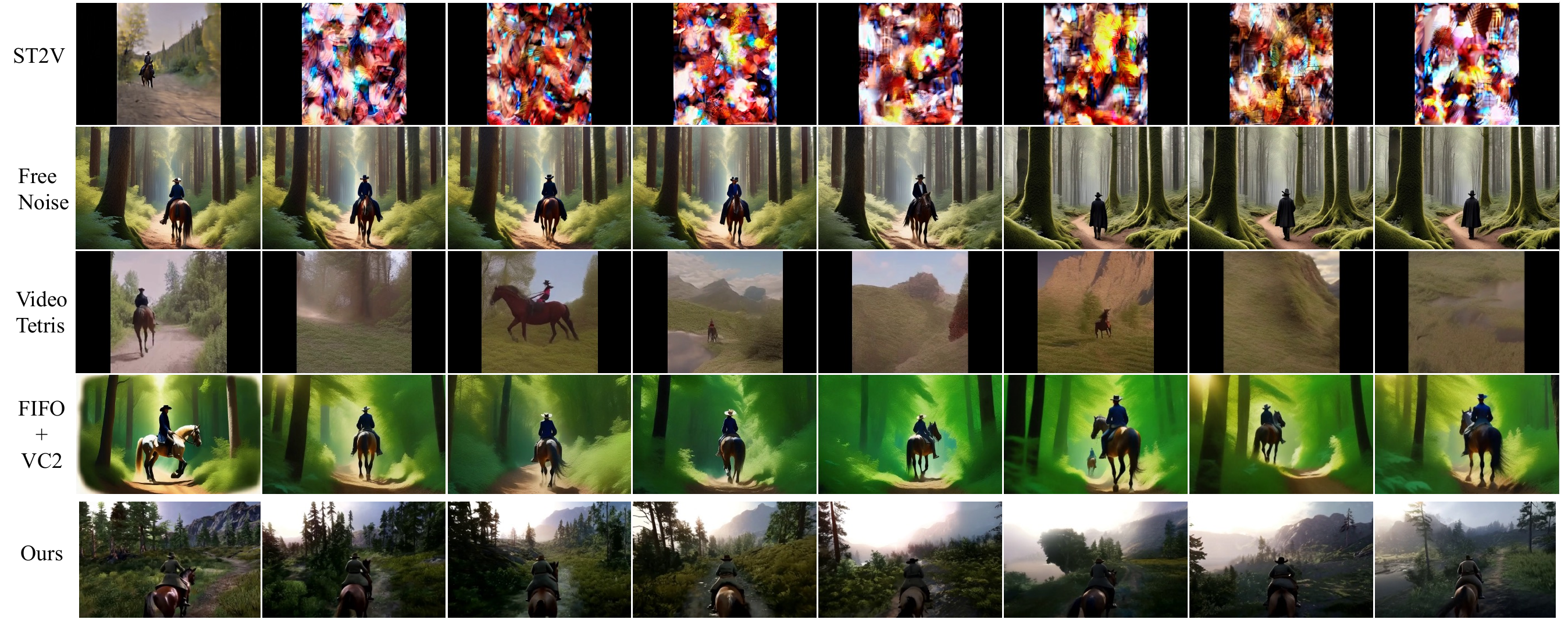}
   \caption{The qualitative comparison. We recommend readers refer to our webpage for video comparisons.}
   \label{fig:comparison}
\end{figure*}

Our expanded comparison includes more baseline methods evaluated with our standard settings, including StreamingT2V \cite{henschel2024streamingt2v}, FreeNoise \cite{qiu2023freenoise}, VideoTetris \cite{tian2024videotetris}, and FIFO-VC2 \cite{kim2024fifo}, as shown in \cref{fig:comparison}. StreamingT2V fails on longer videos, FreeNoise/FIFO+VC2 shows limited dynamics (static subjects), and VideoTetris has rich but illogical variations.

\setlength{\textfloatsep}{1mm}
\begin{table*}[!htbp]
\centering
\setlength{\tabcolsep}{10pt}
\caption{Quantitative evaluation of comparison study.}
\label{tab:comparison}
\resizebox{1.0\linewidth}{!}{
\begin{tabular}{ccccccc|cccc|cc}
\hline
                               & \multicolumn{6}{c|}{VBench}                                                                                                                                                                                                                & \multicolumn{4}{c|}{VBench-Long}                                                                                                                                                                   & \multicolumn{2}{c}{Human Study}                                   \\ \hline
Models                         & SC                            & BC                                                   & TF                            & MS                            & IQ                            & DD                                                  & SC                                                   & BC                                                   & MS                            & DD                                                   & TA                              & MC                              \\ \hline
Video-Inf                      & 81.80                         & 90.56                                                & \cellcolor[HTML]{D3F3FC}96.66 & 97.65                         & 61.58                         & \cellcolor[HTML]{FFCE93}{31.0} & 91.73                                                & 95.63                         & 97.67                         & \cellcolor[HTML]{FFCE93}{28.00} & 0.31\%                          & 0.93\%                          \\
DiTCtrl                        & \cellcolor[HTML]{F8DFC1}76.76 & \cellcolor[HTML]{F8DFC1}{87.96} & 95.52                         & \cellcolor[HTML]{D3F3FC}97.78 & 59.26                         & 75.0                                                & 91.67                                                & 94.21                         & 97.88                         & 53.88                                                & 5.3\%                           & 4.98\%                          \\
ST2V                           & \cellcolor[HTML]{F8A102}67.71 & \cellcolor[HTML]{F8A102}{85.18} & \cellcolor[HTML]{F8A102}93.51 & 94.40                         & 42.53                         & \cellcolor[HTML]{F8DFC1}{34.0} & \cellcolor[HTML]{F8A102}{86.10} & \cellcolor[HTML]{FFCE93}{93.69} & 94.39                         & \cellcolor[HTML]{FFCE93}{25.42} & 0.93\%                          & 0.31\%                          \\
FreeNoise                      & \cellcolor[HTML]{D3F3FC}86.50 & \cellcolor[HTML]{F8F3ED}92.10                        & \cellcolor[HTML]{48D5F8}96.94 & 97.69                         & \cellcolor[HTML]{48D5F8}67.77 & \cellcolor[HTML]{F8A102}{24.0} & \cellcolor[HTML]{48D5F8}96.64                        & \cellcolor[HTML]{48D5F8}{96.52} & \cellcolor[HTML]{D3F3FC}98.02 & \cellcolor[HTML]{F8A102}{18.00} & 1.24\%                          & 2.18\%                          \\
VideoTetris                    & \cellcolor[HTML]{FFCE93}69.27 & \cellcolor[HTML]{FFCE93}{85.86} & \cellcolor[HTML]{FFCE93}94.60 & 97.04                         & 55.95                         & \cellcolor[HTML]{48D5F8}96.0                        & \cellcolor[HTML]{FFCE93}{86.86} & \cellcolor[HTML]{F8A102}{92.84} & 94.73                         & \cellcolor[HTML]{48D5F8}97.12                        & 0.93\%                          & 1.25\%                          \\
FIFO+VC2                       & \cellcolor[HTML]{48D5F8}89.73 & \cellcolor[HTML]{48D5F8}93.93                        & \cellcolor[HTML]{EEF9FC}96.31 & 97.75                         & 60.49                         & \cellcolor[HTML]{F8DFC1}{54.0} & \cellcolor[HTML]{D3F3FC}94.82                        & \cellcolor[HTML]{D3F3FC}{96.43} & 97.79                         & \cellcolor[HTML]{F8DFC1}{49.08} & 4.36\%                          & 3.12\%                          \\
FIFO+CogX                      & 86.22                         & \cellcolor[HTML]{D3F3FC}92.89                        & 94.78                         & 97.48                         & \cellcolor[HTML]{D3F3FC}64.10 & 78.57                                               & 93.78                                                & 95.42                         & 97.43                         & 66.53                                                & \cellcolor[HTML]{D3F3FC}23.36\% & \cellcolor[HTML]{D3F3FC}20.87\% \\ \hline
Ours                           & 84.57                         & 92.21                                                & 95.41                         & \cellcolor[HTML]{48D5F8}98.08 & 63.31                         & \cellcolor[HTML]{D3F3FC}78.95                       & 94.20                                                & 95.52                                                & \cellcolor[HTML]{48D5F8}98.40 & \cellcolor[HTML]{D3F3FC}68.58                        & \cellcolor[HTML]{48D5F8}63.57\% & \cellcolor[HTML]{48D5F8}66.36\% \\
{\color[HTML]{6665CD} TestSet} & {\color[HTML]{6665CD} 85.49}  & {\color[HTML]{6665CD} 91.43}                         & {\color[HTML]{6665CD} 95.62}  & {\color[HTML]{6665CD} 98.33}  & {\color[HTML]{6665CD} 62.78}  & {\color[HTML]{6665CD} 89.00}                        & {\color[HTML]{6665CD} 94.34}                         & {\color[HTML]{6665CD} 95.03}                         & {\color[HTML]{6665CD} 98.35}  & {\color[HTML]{6665CD} 82.50}                         & --                              & --                              \\ \hline
\end{tabular}}
\end{table*}
\definecolor{FIRST}{HTML}{48D5F8}
\definecolor{SECOND}{HTML}{D3F3FC}
\definecolor{subpar}{HTML}{FFCE93}
For quantitative evaluation of added baselines, we use our paper's setup, including a 26-participant human study (\cref{tab:comparison}: \colorbox{FIRST}{first}, \colorbox{SECOND}{second}, \colorbox{subpar}{subpar}). We find that Subject and Background Consistency (\textbf{SC} \& \textbf{BC}) and Temporal Flickering (\textbf{TF}) favor less dynamic videos, \eg, FreeNoise/FIFO+VC2 ranks high in these but low in Dynamic Degree (\textbf{DD}). To further support this, we compute these metrics on MiraData's filtered test set, which features high-quality, continuous motion videos (\textcolor[HTML]{6665CD}{bottom row}). Some methods outperform TestSet on \textbf{SC}, \textbf{BC}, and \textbf{TF}, yet still significantly trail in \textbf{DD}. VideoTetris, with the highest \textbf{DD}, conversely shows lower \textbf{SC} \& \textbf{BC}, indicating potentially disordered, abrupt motions. CogVideoX \cite{yang2024cogvideox} and VBench++ \cite{huang2024vbench++} also report these metric limitations, as \textbf{SC}, \textbf{BC}, and \textbf{TF} assess quality based on neighboring frame similarity (DINO, CLIP, Mean Absolute Error), thus favoring static videos with higher inter-frame similarity. Therefore, reliable quality assessment requires considering both dynamic aspects and these consistency metrics, as also noted by VBench++. Recognizing these limitations, we also evaluate on VBench-Long, a benchmark for long-term consistency that analyzes keyframe similarity across video segments, overcoming the local metrics limitations. Filtering out methods with subpar \textbf{DD} and \textbf{SC}/\textbf{BC}, our method surpasses FIFO+CogX on all four metrics and all other baselines in human evaluations. The evaluation of long video generation quality is still a significant challenge that we will explore further in the future.

\section{Limitations and Discussions}
\label{supp:limitations}
\begin{figure*}[!htbp]
    \centering
    \includegraphics[width=1.0\textwidth]{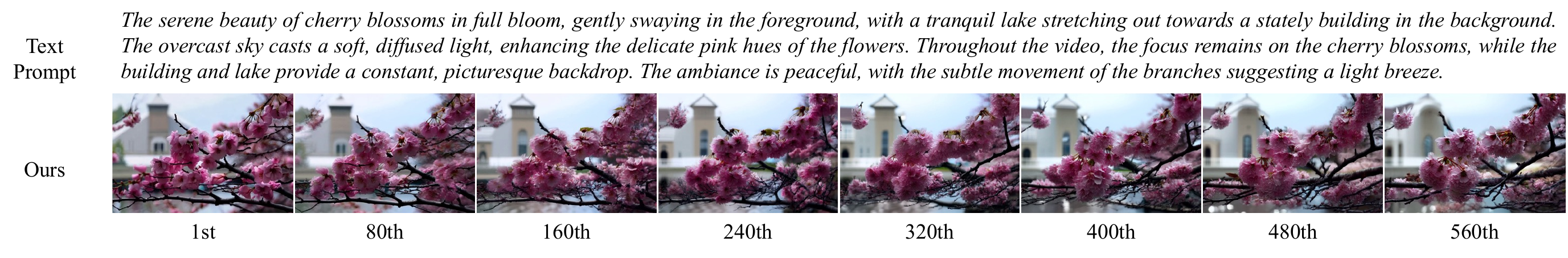}
    \caption{Gradual variations in foreground or background objects over extended sequences.}
    \vspace{0.3cm}
    \label{fig:failure_1}
\end{figure*}

Despite the effectiveness of TokensGen in maintaining long-range consistency, it does not preserve all fine-grained details. Focusing on high-level semantics, tokens may cause gradual variations in foreground or background objects over extended sequences, as shown in \cref{fig:failure_1,fig:editing}.

\begin{figure*}[!htbp]
    \centering
    \includegraphics[width=1.0\textwidth]{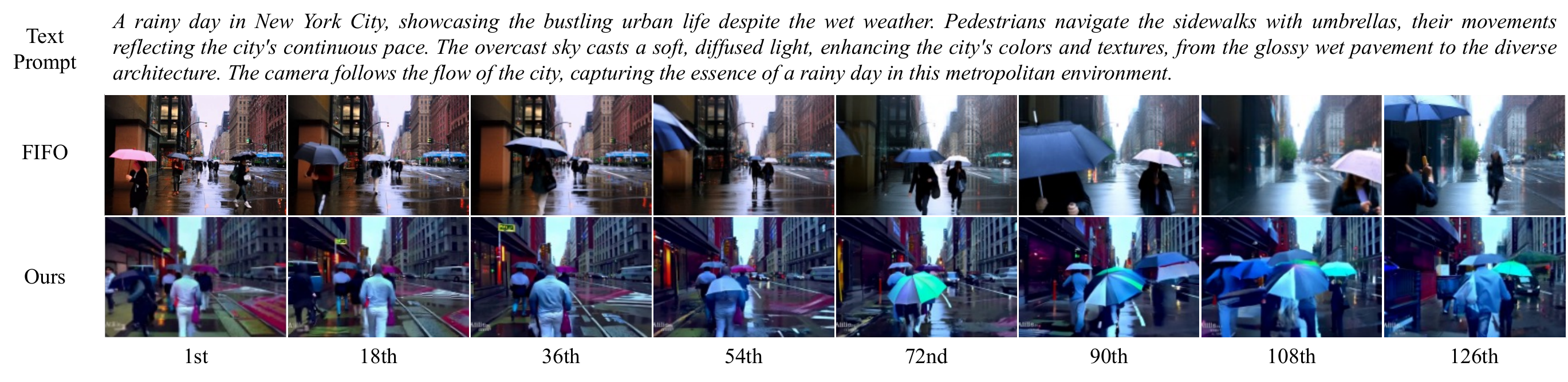}
    \caption{Both the FIFO strategy and the condensed tokens are insufficient to capture intricate spatial-temporal cues, leading to performance limitations.}
    \vspace{0.3cm}
    \label{fig:failure_2}
\end{figure*}

Our current framework employs a tuning-free FIFO strategy to maintain short-term consistency during inference. While effective in many scenarios, FIFO can deliver suboptimal performance for cross-clip temporal consistency in some complex scenes. In such cases, the condensed tokens are also insufficient to capture intricate spatial-temporal cues, leading to performance limitations. We illustrate these failure cases in \cref{fig:failure_2}. Addressing these challenges will require more fine-grained tokenization and stronger short-term consistency strategies beyond tuning-free FIFO. 


Our framework is trained and tested on a limited dataset of gameplay and landscape videos, but is scalable to larger datasets for broader applications. In future work, exploring multi-scale tokenization or hybrid representations could bolster fine-grained controllability, retaining subtle attributes while preserving the scalability and resource efficiency.

\section{Additional Visual Results}
For more visual results, comparisons, and ablation studies, please refer to our webpage.
\label{supp:add_results}

\end{document}